\pdfoutput=1

\documentclass[11pt,a4paper]{article}
\usepackage[hyperref]{acl2021}
\usepackage{times}
\usepackage{latexsym}

\usepackage{microtype}

\usepackage{graphicx}
\usepackage{subcaption}
\usepackage{amsmath}
\usepackage{amssymb}
\usepackage{booktabs}
\usepackage{multirow}
\usepackage{enumitem}
\usepackage{cleveref}

\usepackage{changepage}

\usepackage{pifont}% http://ctan.org/pkg/pifont
\aclfinalcopy % Uncomment this line for the final submission

\title{A Few More Examples May Be Worth Billions of Parameters}

\author{
  Yuval Kirstain$^{\spadesuit{}}$ \hspace{0.15cm}
  Patrick Lewis$^{\dagger{}\ddagger{}}$ \hspace{0.15cm}
  Sebastian Riedel$^{\dagger{}\ddagger{}}$ \hspace{0.15cm}
  Omer Levy$^{\spadesuit{}\ddagger{}}$ \\
$^\spadesuit$ Tel-Aviv University \\    
  \hspace{0.3cm} $^{\dagger{}}$ University College London \\ 
  $^{\ddagger{}}$ Facebook AI Research \\
  \small{\texttt{\{yuval.kirstain,levyomer\}@cs.tau.ac.il}},  \small{\texttt{\{patrick.lewis,s.riedel\}@cs.ucl.ac.uk}}
}

\date{}

\begin{document}
\maketitle
\begin{abstract}
We investigate the dynamics of increasing the number of model parameters versus the number of labeled examples across a wide variety of tasks.
Our exploration reveals that while scaling parameters consistently yields performance improvements, the contribution of additional examples highly depends on the task's format.
Specifically, in open question answering tasks, enlarging the training set does not improve performance.
In contrast, classification, extractive question answering, and multiple choice tasks benefit so much from additional examples that collecting a few hundred examples is often ``worth'' billions of parameters.
We hypothesize that unlike open question answering, which involves recalling specific information, solving strategies for tasks with a more restricted output space transfer across examples, and can therefore be learned with small amounts of labeled data.\footnote{Our code is publicly available: \url{https://github.com/yuvalkirstain/lm-evaluation-harness}.}
\end{abstract}
\section{Introduction}

Recent work on few-shot learning for natural language tasks explores the dynamics of scaling up either the number of model parameters~\cite{brown2020language} or labeled examples~\cite{le-scao-rush-2021-many}, while controlling for the other variable by setting it to a constant.
%, which might not necessarily reflect a realistic few-shot setting.
For example, \citet{brown2020language} focus on in-context learning from roughly 32 to 64 examples, a practice that was adopted by fine-tuning approaches as well~\cite{schick-schutze-2021-just,gao-etal-2021-making,Tam2021ImprovingAS}; however, there are many practical few-shot scenarios where \textit{hundreds} of examples can be collected at a relatively low effort.\footnote{In SQuAD~\cite{rajpurkar-etal-2016-squad}, for example, the average annotation pace is around one minute per question, producing 480 examples in a single 8-hour workday.}
Other work experiments with single-size models~\cite{Schick2020FewShotTG,ram-etal-2021-shot,le-scao-rush-2021-many,gao-etal-2021-making}, even though larger (or smaller) models may exhibit different behavior.
Furthermore, much of the literature focuses on classification tasks \cite{schick-schutze-2021-exploiting,gao-etal-2021-making,le-scao-rush-2021-many}, leaving it unclear whether their conclusions generalize to tasks with less restricted output spaces.

\begin{figure}[t!]
\centering
\subfloat{\includegraphics[width=0.45\linewidth]{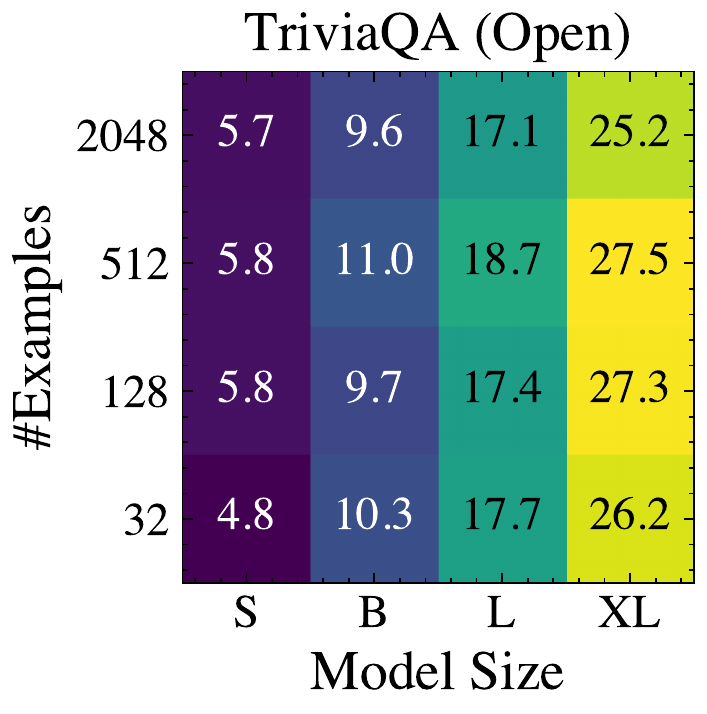}}\qquad
\subfloat{\includegraphics[width=0.45\linewidth]{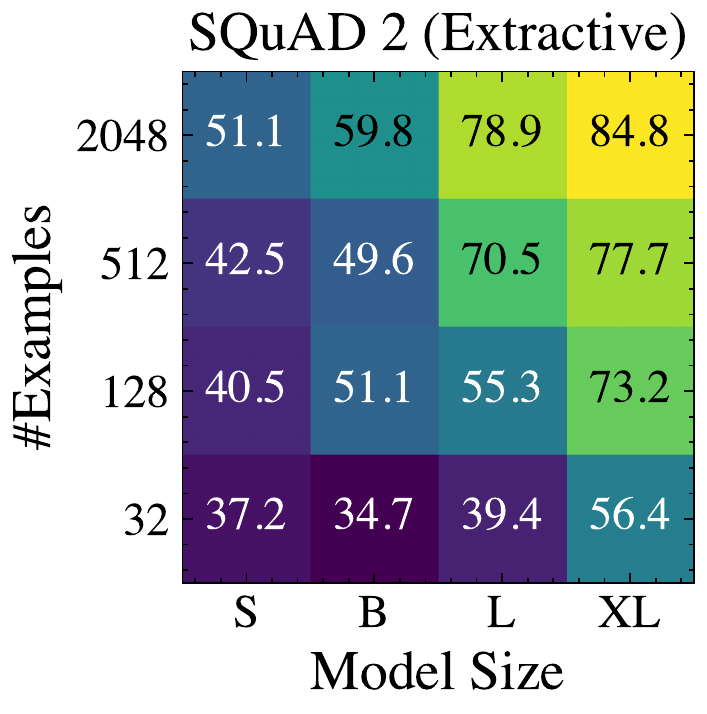}}
\caption{Open QA tasks (e.g. TriviaQA) benefit from additional parameters exclusively, while extractive QA tasks (e.g. SQuAD 2) benefit from both larger models and more labeled data.}
\label{fig:first}
\end{figure}

In this paper, we conduct a systematic exploration of few-shot learning for language tasks, where we investigate the dynamics of increasing the number of model parameters (using different sizes of the self-supervised T5 \cite{2020t5}) versus the number of target-task labeled examples (from 32 to 2048) across a variety of tasks, including not only classification, but also extractive, multiple-choice, and open question answering.
Overall, we evaluate 192 scenarios by training 7,680 models to control for hyperparameters and random seeds.

Our experiments show that, surprisingly, the contribution of additional parameters versus additional labeled examples highly depends on the \textit{format} of the task.
For open QA tasks, such as the open-domain version of Natural Questions~\cite{kwiatkowski-etal-2019-natural, lee-etal-2019-latent}, 
which require the model to recall specific information seen during pretraining,
\textit{enlarging the training set does not improve performance}.
By contrast, increasing the number of model parameters results in substantial gains (see TriviaQA~\cite{joshi-etal-2017-triviaqa} in \Cref{fig:first}). 
Hence, when dealing with open QA, model parameters are of immense value, and cannot be replaced by increasing the number of labeled examples.

On the other hand, we observe a completely different trend for classification, extractive QA, and multiple-choice tasks.
These tasks benefit from enlarging \textit{both} the training set and the model (see SQuAD 2~\cite{rajpurkar-etal-2018-know} in \Cref{fig:first}).
We observe that hundreds of examples are often ``worth'' \textit{billions} of parameters;
T5-L fine-tuned on 4 times more data is roughly competitive with T5-XL, which has 4 times the number of parameters.
Moreover, some tasks benefit so much from labeled examples, that collecting even 512 data points can make a fine-tuned T5-L (800M parameters) outperform GPT-3 (175B parameters).

Finally, we hypothesize that unlike open QA, formats with restricted output spaces have solving strategies (such as elimination) that can be learned from small amounts of labeled data.
This hypothesis also provides a possible explanation as to why lean retrieve-and-read approaches (such as DrQA~\cite{chen-etal-2017-reading}, ORQA \cite{lee-etal-2019-latent}, and DPR~\cite{karpukhin-etal-2020-dense})) appear to be more robust than multi-billion-parameter closed-book models \cite{roberts-etal-2020-much} when tested on non-overlapping data \cite{lewis-etal-2021-question}.

\section{Experiments}
\label{sec:experiments}

We describe the tasks (Section~\ref{sec:tasks}), models (Section~\ref{sec:models}), data regimes (Section~\ref{sec:data}), and implementation details (Section~\ref{sec:setup}) of our systematic experiment suite.
In total, we experiment with 12 tasks, 4 models, 4 data regimes (with 5 samples each), and 8 hyperparameter configurations; these amount to 7,680 trained models, evaluated across 192 task-model-data scenarios.

\subsection{Datasets}
\label{sec:tasks}

We experiment with 12 datasets, divided into 4 broad types of \emph{task formats}. The formats and their constituent tasks are described below.

\paragraph{Classification}
In classification tasks, the model is expected to read a given text and predict a single label from a small closed set, e.g. \textit{yes} or \textit{no}.
We adopt classification tasks from the GLUE~\cite{wang-etal-2018-glue} and SuperGLUE~\cite{Wang2019SuperGLUEAS} benchmarks, namely:
Recognizing Textual Entailment~\cite[RTE,][]{Dagan-RTE, BarHaim2006TheSP, Giampiccolo2007TheTP, Bentivogli2009TheSP},
the Stanford Sentiment Treebank~\cite[SST-2,][]{socher-etal-2013-recursive},
and BoolQ~\cite{clark-etal-2019-boolq}. We report accuracy for all classification datasets.

\paragraph{Extractive Question Answering}
In extractive QA, the model is given a passage and a question, and is then expected to produce an answer in the form of a span from the passage.
We experiment with SQuAD 2~\cite{rajpurkar-etal-2018-know}, 
HotpotQA~\cite{yang-etal-2018-hotpotqa},
and DROP~\cite{dua-etal-2019-drop}.
Each of these datasets contains an additional ``quirk'' that makes it more challenging than the original SQuAD dataset~\cite{rajpurkar-etal-2016-squad}, which popularized the extractive QA format:
SQuAD 2 has unanswerable questions, HotpotQA provides multiple passages per question, and DROP contains many arithmetic questions whose answer is not strictly extractive, but can be derived from a set of spans in the given passage. For all extractive QA datasets we report token-wise F1.

\paragraph{Multiple Choice}
Multiple choice tasks provide the model with a question and several candidate answers, with the goal of selecting the correct one.
We focus on three datasets in this format:
the easy question set from the AI2 Reasoning Challenge~\cite[ARC-E,][]{Clark2018ThinkYH},
the Physical Interaction Question Answering dataset~\cite[PIQA,][]{Bisk2020PIQARA},
and CommonsenseQA~\cite{talmor-etal-2019-commonsenseqa}.
Unlike extractive QA, multiple choice tasks do \textit{not} contain supporting evidence (a passage) for answering the question, and in contrast to classification, they have a different output space (candidate answers) for each example. We report accuracy for all multiple choice datasets.

\paragraph{Open Question Answering}
Open QA\footnote{We deviate from the widely-used term \textit{open-domain QA}, which describes the task, and use \textit{open QA} instead to refer to the format, much like we use \textit{extractive QA} rather than \textit{reading comprehension}.}
datasets provide the model with just a question; no supporting evidence or closed candidate set is available.
We experiment with open-domain versions of Natural Questions~\cite{kwiatkowski-etal-2019-natural}, TriviaQA~\cite{joshi-etal-2017-triviaqa}, and SQuAD 1~\cite{rajpurkar-etal-2016-squad}.
Our experiments here focus purely on the closed-book setting~\cite{roberts-etal-2020-much}, which does not allow models to retrieve text from an external corpus, restricting them to information stored in their parameters. For all open QA tasks we report F1 as the main metric.

\subsection{Models}
\label{sec:models}

The Text-to-Text Transfer Transformer~\cite[T5,][]{2020t5} uses an encoder-decoder transformer architecture. 
It is pretrained on the task of generating masked-out spans over the Colossal Clean Crawled Corpus (C4), which contains 800GB of English-language text.
We use version 1.1 of T5, which is not trained on any labeled data.
Our experiments include the 77M (S), 250M (B), 800M (L), and 3B (XL) parameter variants of this model. 

\subsection{Training Data}
\label{sec:data}
While many publicly released datasets include an enormous number of labeled examples~\cite{rajpurkar-etal-2018-know, yang-etal-2018-hotpotqa, kwiatkowski-etal-2019-natural}, and recent work on few-shot learning focus on an extreme scenario in which less than one hundred examples are at hand~\cite{brown2020language, schick-schutze-2021-exploiting, gao-etal-2021-making}, we choose to simulate a broader set of practical scenarios where a small-to-medium dataset is available; in SQuAD~\cite{rajpurkar-etal-2016-squad}, for example, the average annotation pace is around one minute per question, producing 480 examples in a single 8-hour workday.
Therefore, we consider 4 dataset sizes for each task: 32, 128, 512, and 2048 examples. 
For each dataset size, we sample the relevant amount of examples five times using different seeds, thus creating 20 datasets for each task overall.
We report the average score for each dataset size, thereby reducing the high variance associated with training on small datasets.

\subsection{Implementation}
\label{sec:setup}

\paragraph{Code}
For our implementation we extend EleutherAI's language model evaluation harness~\cite{lm-evaluation-harness} to allow fine-tuning and evaluating additional datasets and models.

\paragraph{Prompts}
We adopt the prompts used by \citet{brown2020language} and \citet{khashabi-etal-2020-unifiedqa}, with minimal adaptations to T5 by adding a mask token followed by a period. 
Following \citet{le-scao-rush-2021-many}, we use prompts in conjunction with fine-tuning. 
% Unlike \citet{brown2020language}, we do not prepend questions and answers (other than the target question) that belong to the same context as demonstrations, as this scenario is unrealistic.

\paragraph{Decoding}
We use greedy decoding for extractive and open QA tasks.
For classification and multiple choice tasks, we compare the model's probability for each possible outcome,
and predict the option with the highest probability.
In BoolQ, for example, we compare $P(\text{``yes''} | x)$ (the probability of the positive class) with $P(\text{``no''} | x)$ (the probability of the negative class), where $x$ is the prompt containing the context and the question.

\paragraph{Hyperparameters}
To tune hyperparameters for fine-tuning, we split the available data into 75\% training and 25\% validation (e.g. 24 training examples and 8 validation when the dataset size is 32).
For each case, we experiment with two learning rates ($5\mathrm{e}{-5}$, $5\mathrm{e}{-4}$) combined with linear decay, two weight decays ($0.001$, $0.1$), and two values for amount of steps (512, 2048).
The effective batch size is always 32 examples.
Additionally, we use a dropout ratio of 0.1, gradient clipping is set to 1, and the amount of warmup steps is determined by the maximum between 10\% of the training steps and 100.
We evaluate each run after every epoch and choose the model with the lowest validation loss for extractive and open QA tasks, and highest validation accuracy for classification and multiple choice tasks.

\section{Results}
\label{sec:results}

\begin{figure*}[t!]
\centering
\small
\begin{tabular}{@{}cccc@{}}
\toprule
\textbf{~~~~~Classification} & \textbf{~~~~~Extractive QA} & \textbf{~~~~~Multiple Choice} & \textbf{~~~~~Open QA} \\
\midrule
\subfloat{\includegraphics[width=0.45\columnwidth]{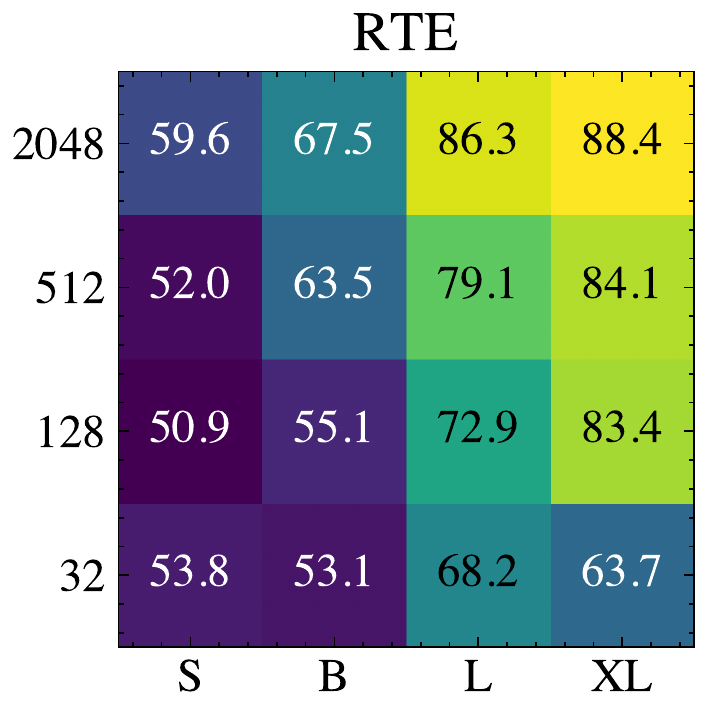}} & 
\subfloat{\includegraphics[width=0.45\columnwidth]{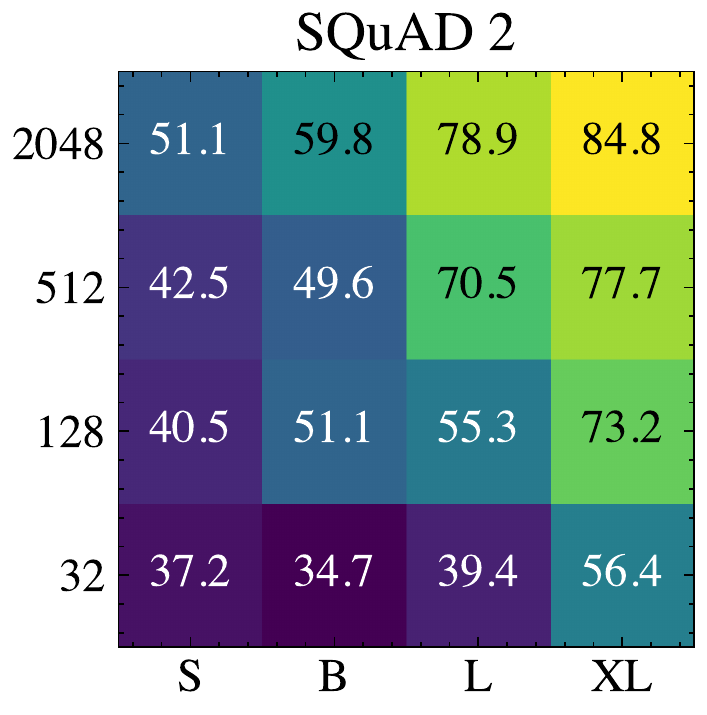}} & 
\subfloat{\includegraphics[width=0.45\columnwidth]{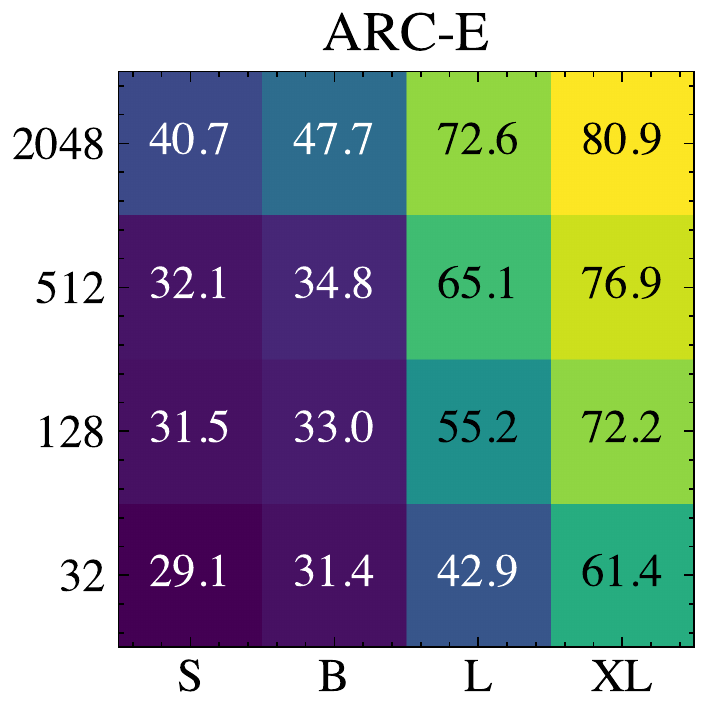}} &
\subfloat{\includegraphics[width=0.45\columnwidth]{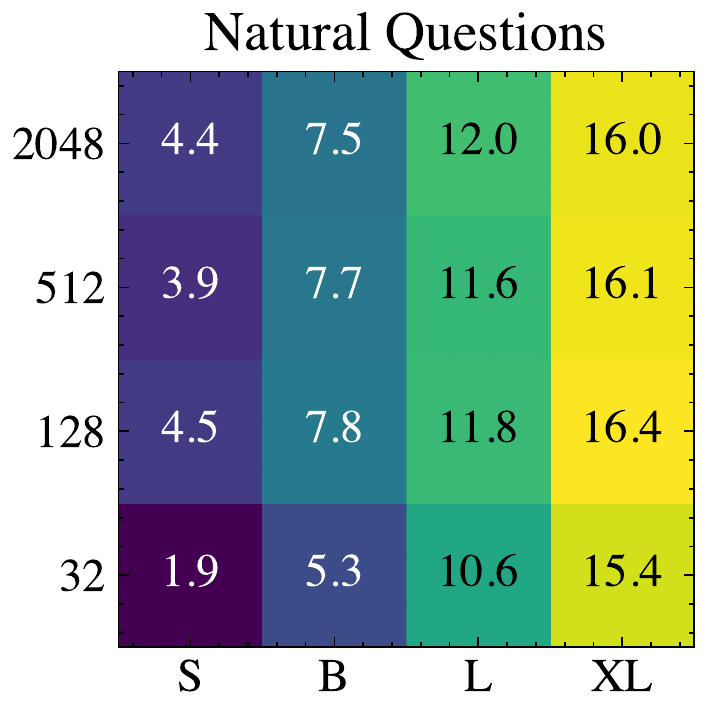}} \\
\subfloat{\includegraphics[width=0.45\columnwidth]{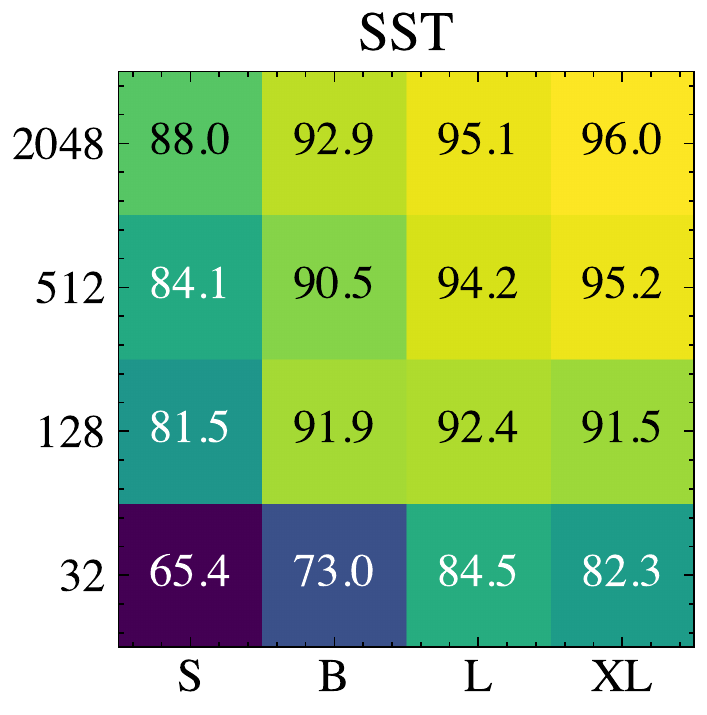}} &
\subfloat{\includegraphics[width=0.45\columnwidth]{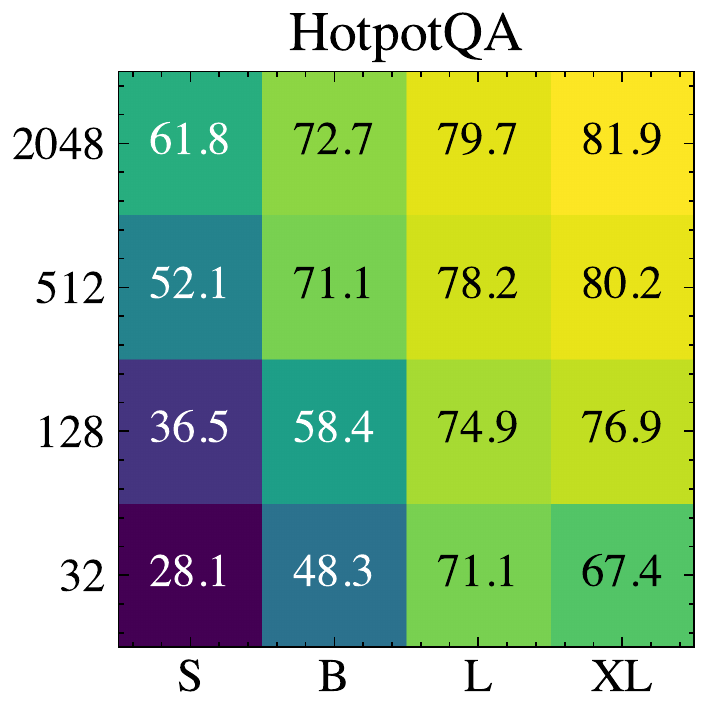}} &
\subfloat{\includegraphics[width=0.45\columnwidth]{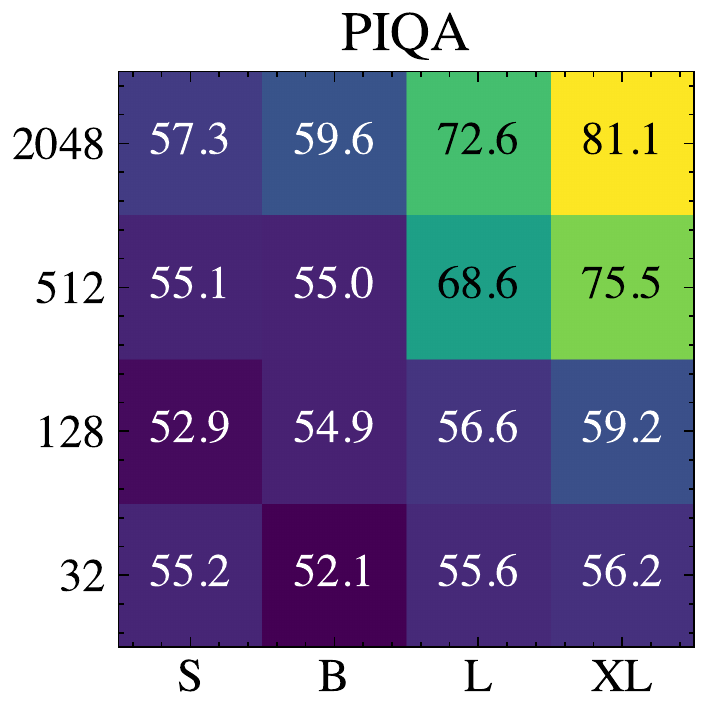}} &
\subfloat{\includegraphics[width=0.45\columnwidth]{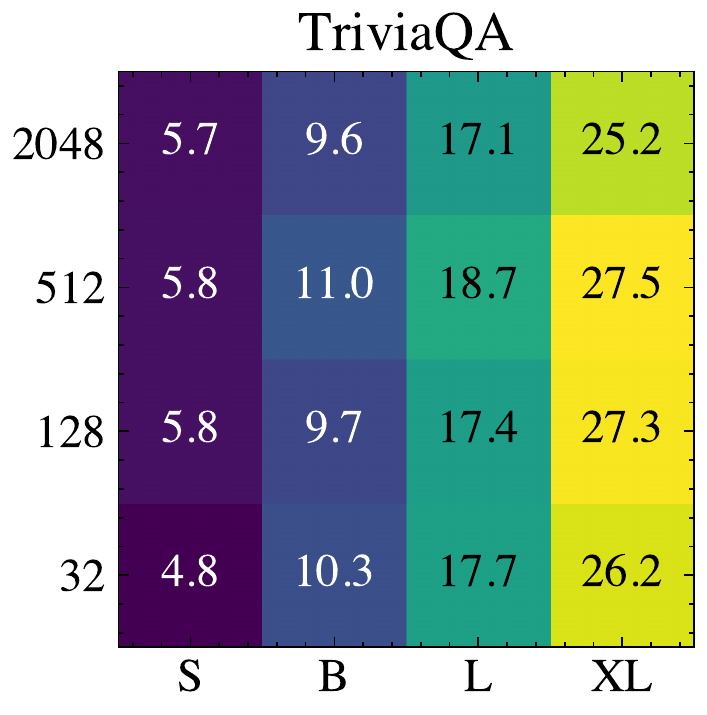}} \\
\subfloat{\includegraphics[width=0.45\columnwidth]{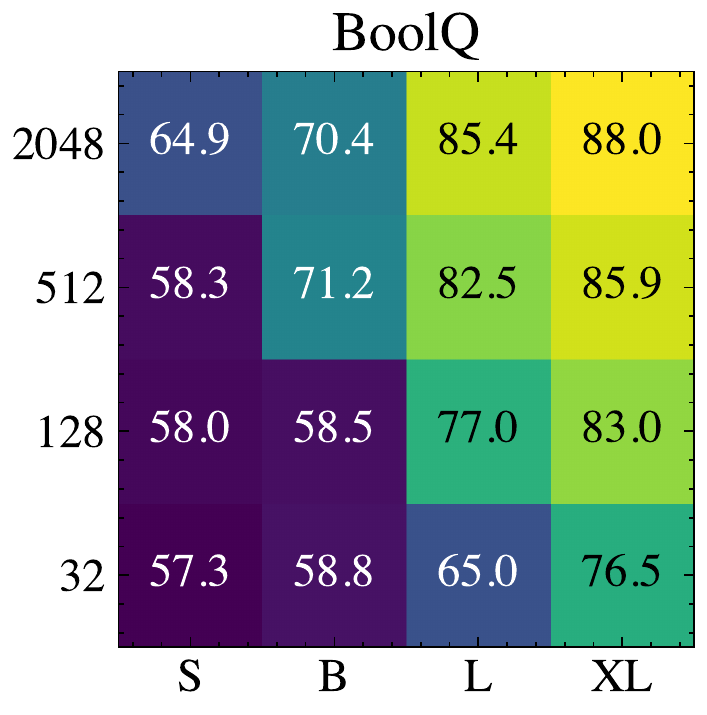}} &
\subfloat{\includegraphics[width=0.45\columnwidth]{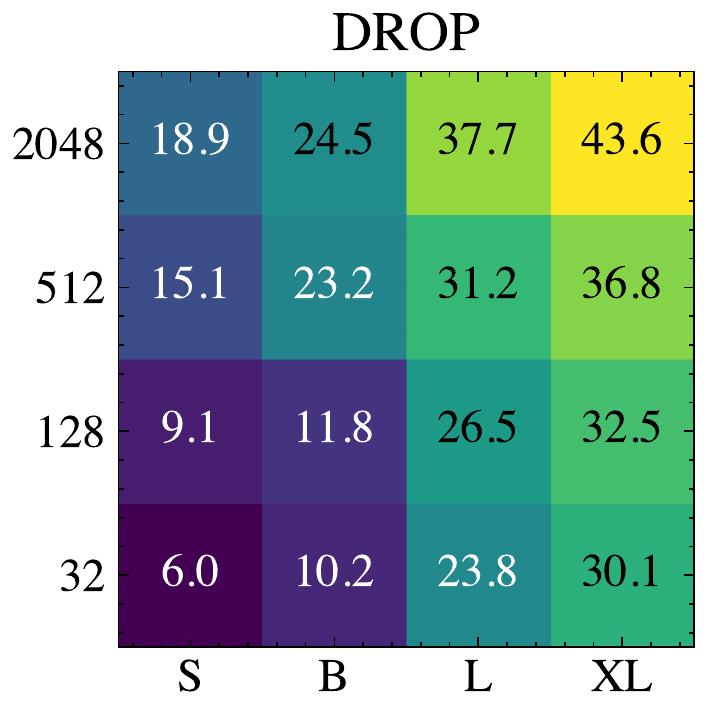}} &
\subfloat{\includegraphics[width=0.45\columnwidth]{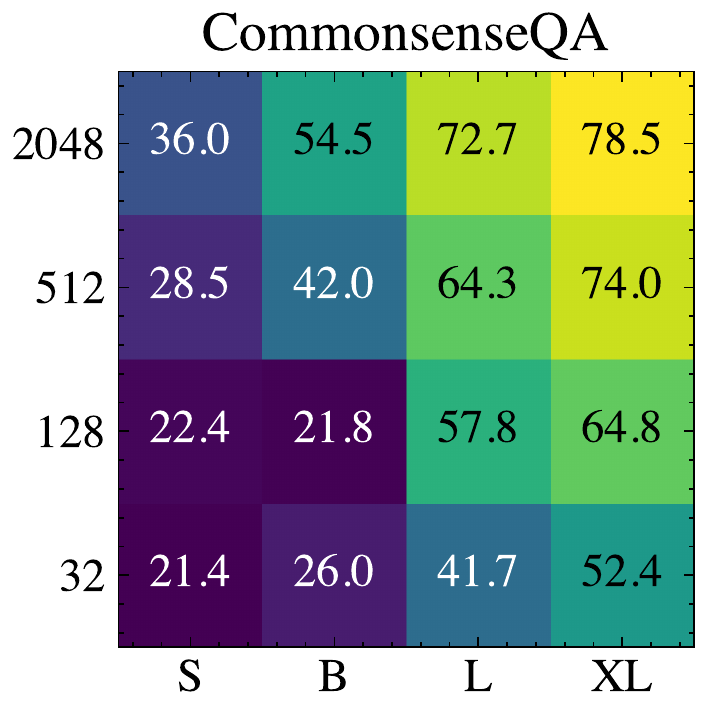}} &
\subfloat{\includegraphics[width=0.45\columnwidth]{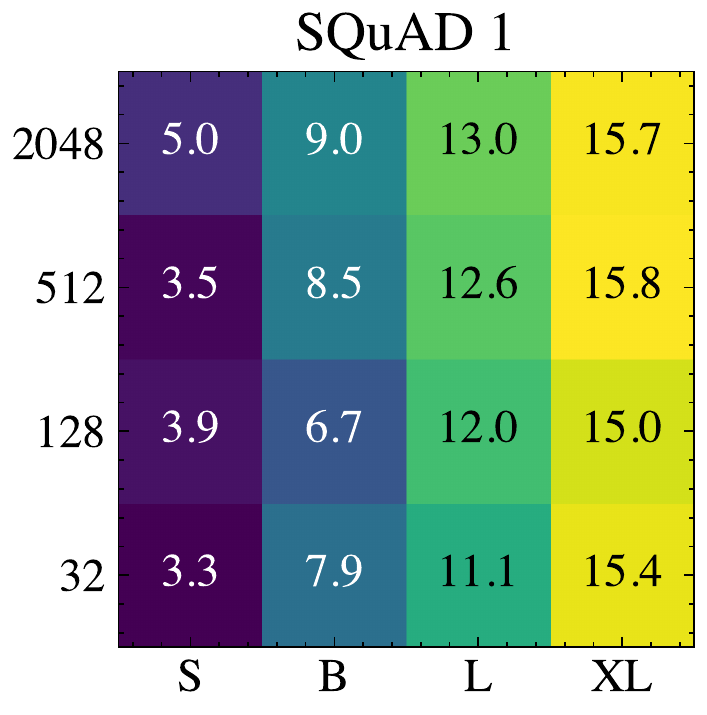}} \\
\bottomrule
\end{tabular}
\caption{Each heatmap displays the model's performance (F1/accuracy) given its size in parameters (horizontal axis) and the number of labeled examples available during fine-tuning (vertical axis).}
\label{fig:main_heatmaps}
\end{figure*}

\begin{figure}[t!]
\centering
\small
\begin{tabular}{@{}cc@{}}
\toprule
 \textbf{~~~~~Multiple Choice} & \textbf{~~~~~Open QA} \\
\midrule
\subfloat{\includegraphics[width=0.45\columnwidth]{heatmaps/heat_map_arc_easy_extractive.pdf}} &
\subfloat{\includegraphics[width=0.45\columnwidth]{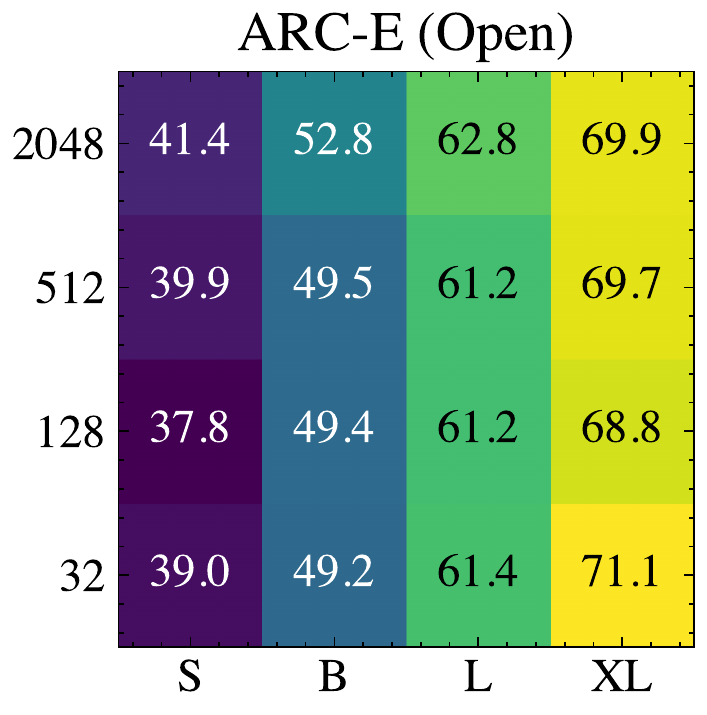}} \\
\subfloat{\includegraphics[width=0.45\columnwidth]{heatmaps/heat_map_piqa_extractive.pdf}} &
\subfloat{\includegraphics[width=0.45\columnwidth]{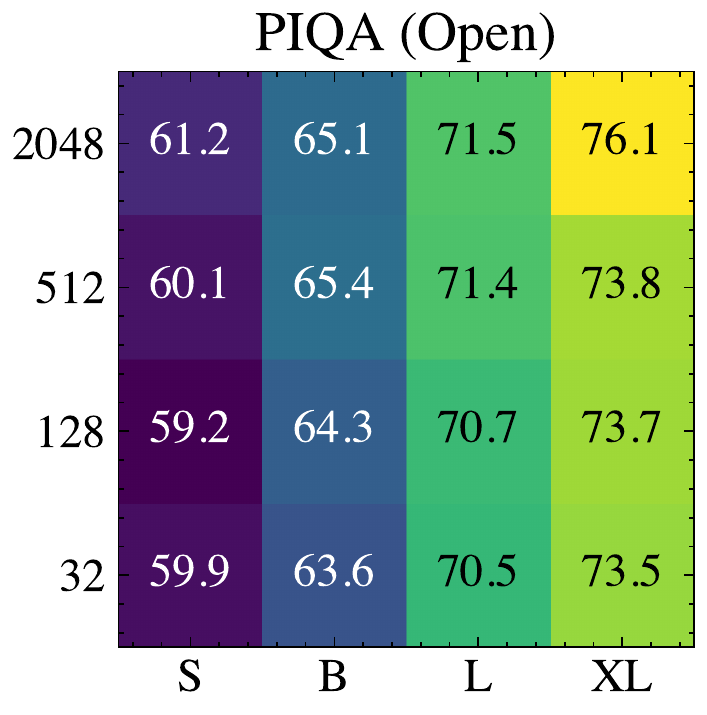}} \\
\subfloat{\includegraphics[width=0.45\columnwidth]{heatmaps/heat_map_commonsense_qa_extractive.pdf}} &
\subfloat{\includegraphics[width=0.45\columnwidth]{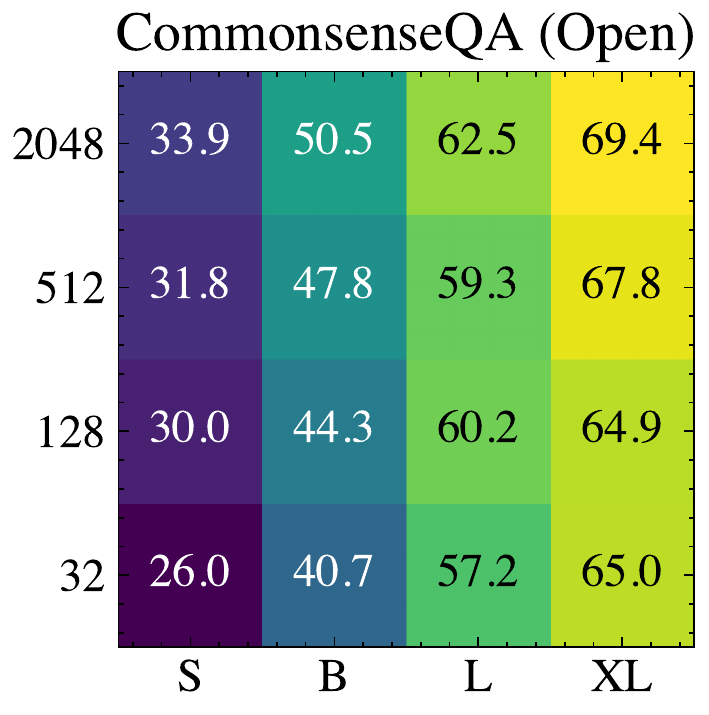}} \\
\bottomrule
\end{tabular}
\caption{Converting multiple choice tasks (left column) to open QA (right column) changes the scaling dynamics, replacing diagonal gradients (performance improves with more parameters and more data) to horizontal gradients (performance improves almost exclusively with more parameters).}
\label{fig:mc_vs_open_heatmaps}
\end{figure}

\begin{figure*}[t!]
\centering
\small
\begin{tabular}{@{}ccc@{}}
\toprule
 \textbf{~~~~~Extractive QA} &
 \textbf{~~~~~Multiple Choice} & \textbf{~~~~~Open QA} \\
\midrule
\subfloat{\includegraphics[width=0.45\columnwidth]{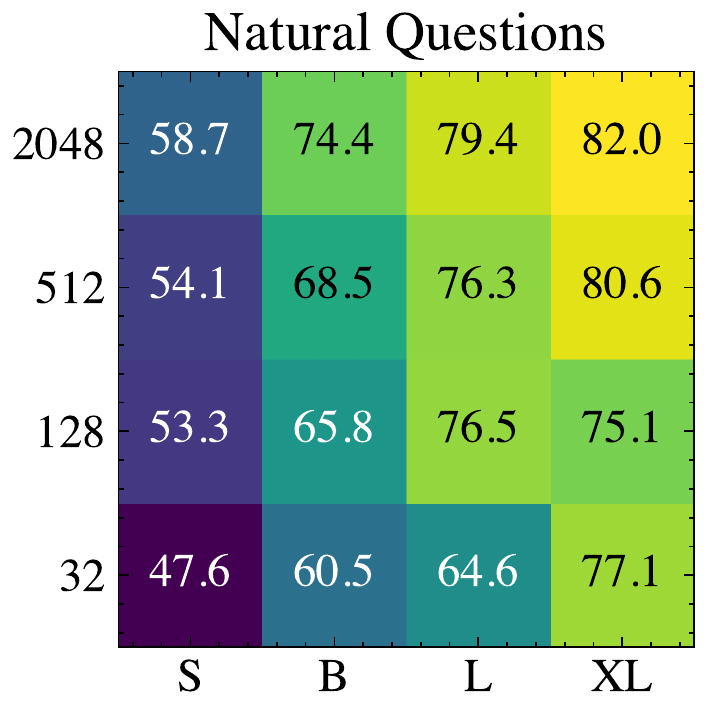}} &
\subfloat{\includegraphics[width=0.45\columnwidth]{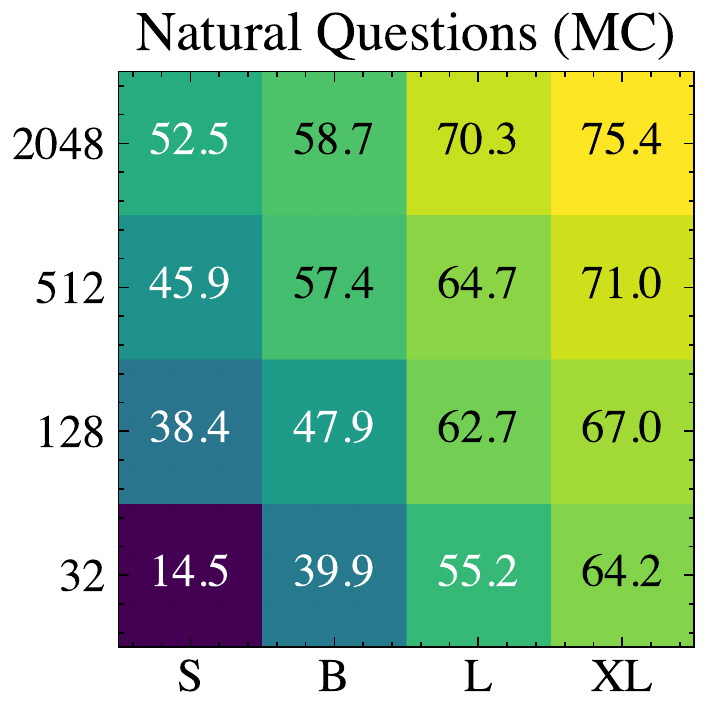}} &
\subfloat{\includegraphics[width=0.45\columnwidth]{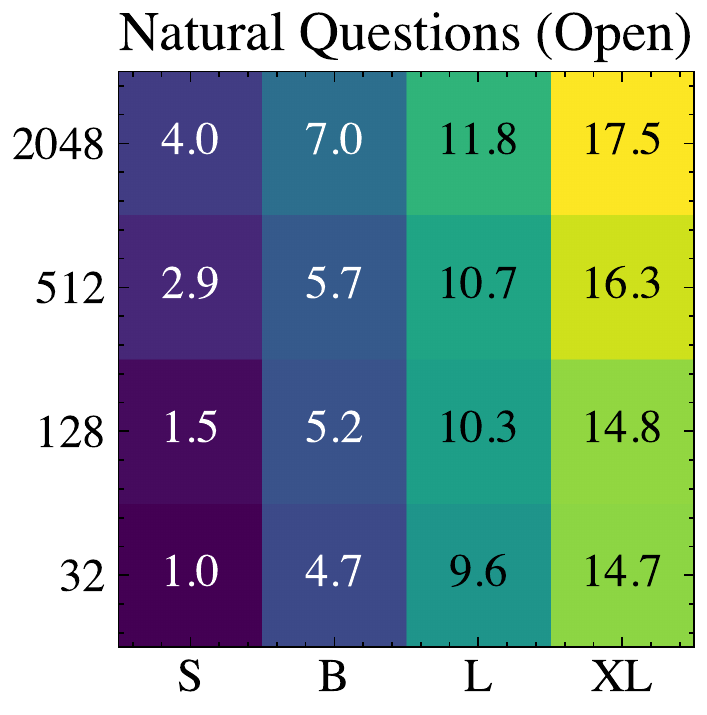}} \\
\bottomrule
\end{tabular}
\caption{Converting Natural Questions \cite{kwiatkowski-etal-2019-natural} from its open QA format (right) to multiple choice (middle) and extractive QA (left) changes the scaling dynamics, replacing horizontal gradients (performance improves almost exclusively with more parameters)
with diagonal gradients (performance improves with more parameters and more data).}
\label{fig:nq_formats_heatmaps}
\end{figure*}

For each task in our experiment suite, we present a heatmap of the model's performance as a function of model size and the number of labeled examples.
These heatmaps expose 
% different trends for two subsets of tasks: whilst most tasks benefit from both larger models and more training data, the performance for open QA tasks \emph{only} improves with additional model parameters (Section~\ref{subsec:main_trends}).
that most tasks benefit from both larger models \textit{and} more training data, to a point where enlarging the dataset will result in similar gains as increasing the number of parameters.
However, this trend does not apply to open QA tasks, whose performance \emph{only} improves with additional model parameters (Section~\ref{subsec:main_trends}). 
Furthermore, we show that converting multiple choice datasets to the open QA format \textit{disables} the benefits of additional training data, whereas converting in the opposite direction -- from open QA to extractive or multiple-choice QA -- \textit{enables} models to improve with more examples (Section~\ref{subsec:format_switch}).
Next, we describe a method for quantifying the relative benefit from parameters versus training data, which confirms the observed trends (Section~\ref{subsec:importance_metric}).
We then show that collecting a few hundred examples allows the much smaller T5-L to outperform GPT3, but not in open QA tasks, where the massive amount of parameters is the prime contributor (Section~\ref{subsec:comparison}).
Finally, we suggest a hypothesis to explain the observed trends (Section~\ref{subsec:discussion}).

\subsection{Main Trends}
\label{subsec:main_trends}

\Cref{fig:main_heatmaps} shows performance as a function of model size and dataset size per task.\footnote{The results are available in tabular form in \Cref{app:full_results}.}
Visualizing the results via heatmaps highlights two patterns:
(1) \textit{diagonal gradients}, where performance significantly improves along both axes (though not necessarily equally), and 
(2) \textit{horizontal gradients}, where performance improves almost exclusively along the horizontal (model size) axis.
We observe that all three open QA datasets exhibit horizontal gradients, while the remaining datasets follow the diagonal patterns.
We do not observe \textit{vertical gradients} at all, indicating that enlarging the model's size is consistently beneficial.

Consider TriviaQA, for example (\Cref{fig:main_heatmaps}, right column, second row); performance approximately doubles when switching models from T5-B to T5-L (and from T5-L to T5-XL), but changes by less than 2 points when increasing the dataset from 32 examples to 2048.
On the other hand, in the classification task SST-2 (\Cref{fig:main_heatmaps}, left column, second row), annotating 128 examples rather than 32 examples results in double-digit improvements for T5-S and T5-B, and in significant gains for larger models as well.
Here, data-driven improvements coincide with parameter-driven improvements, and increasing either factor typically boosts performance.
Moreover, the diagonal gradients show that in many cases a model trained on more data can ``catch up'' with a larger model.
This trend is particularly striking when comparing T5-L with T5-XL, where training the smaller model (T5-L) on 4 times more data is almost always competitive with the larger model (T5-XL).

\subsection{Same Dataset, Different Format}
\label{subsec:format_switch}

While a clear dichotomy arises from Section~\ref{subsec:main_trends} with respect to format, it might also result from the fact that the different datasets were collected and annotated using different methodologies.
Can we conduct a more controlled experiment, which uses the same dataset but in different formats?

We first take the three multiple choice datasets (ARC-E, PIQA, and CommonsenseQA) and convert them into the open QA format by excluding the candidate answers from the input.\footnote{We control for the inference method by selecting the most probable answer candidate, rather than applying greedy decoding. Thus, the only difference between each pair of datasets is whether or not the candidates appear in the input.}
\Cref{fig:mc_vs_open_heatmaps} shows that the diagonal gradients clearly seen in the multiple choice format are replaced with horizontal gradients similar to those of other open QA datasets.%\footnote{The results are available in tabular form in \Cref{app:mc}.}

We also examine data conversion in the opposite direction, by using multiple choice and extractive QA versions of Natural Questions.\footnote{The original Natural Question dataset~\cite{kwiatkowski-etal-2019-natural} is in the extractive QA format; specifically, we use the version in the 2019 MRQA Shared Task~\cite{fisch-etal-2019-mrqa}. We filter the dataset to include only named entity answers that were recognized using an off-the-shelf OntoNotes Named Entity Recognition model from spaCy~\cite{ontonotes, spacy}, and suggest  them as candidate answers alongside entities of the same type that appear in the background passage.} Here we control for the change in format by decoding the multiple choice models as we do for extractive and open QA tasks and report F1.
\Cref{fig:nq_formats_heatmaps} shows that while the open QA heatmap displays largely horizontal gradients, both extractive QA and multiple choice heatmaps follow the diagonal patterns.
Unlike the original open-domain Natural Questions dataset, we do observe some minor improvement along the data axis in this entity-focused version, but analyzing the data reveals that this stems from an increase in example overlap~\cite{lewis-etal-2021-question}, with 11.7\% of test-set answers appearing in the 2048-example training sets, compared to 8.5\% in the original.
Overall, both experiments' results indicate that the task's format directly impacts whether more labeled data will improve performance or not.

\subsection{Quantifying the Relative Impact of Parameters versus Examples}
\label{subsec:importance_metric}

For many tasks, both additional model parameters and labeled examples can improve performance.
However, it is not always clear \textit{how much} each factor contributes to greater performance gains with respect to the other.
To quantify the importance of increasing parameters versus examples, we compute a regression-based metric using the numerical results in a given heatmap.
Specifically, we train the following linear regression model for each heatmap:
\begin{align*}
y = \alpha_m x_m + \alpha_d x_d + b
\end{align*}
where $y$ is the model's performance on the task, $x_m$ is the normalized number of model parameters (S is 1, B is 2, L is 3, and XL is 4), and $x_d$ is the normalized number of dataset examples (32 is 1, 128 is 2, 512 is 3, and 2048 is 4).
The regression coefficients $\alpha_m , \alpha_d$ are scalars, learnt for each task, which are then normalized to measure the relative impact of each axis (parameters versus examples):
\begin{align*}
I_m = \frac{|\alpha_m|}{|\alpha_m| + |\alpha_d|}
\end{align*}
When $0 < I_m < 0.5$, additional examples are greater contributors to performance gains, while $0.5 < I_m < 1.0$ indicates that model parameters have higher relative importance.
% are driving the increase in performance.

\begin{figure}[t!]
\centering
\includegraphics[width=1\linewidth]{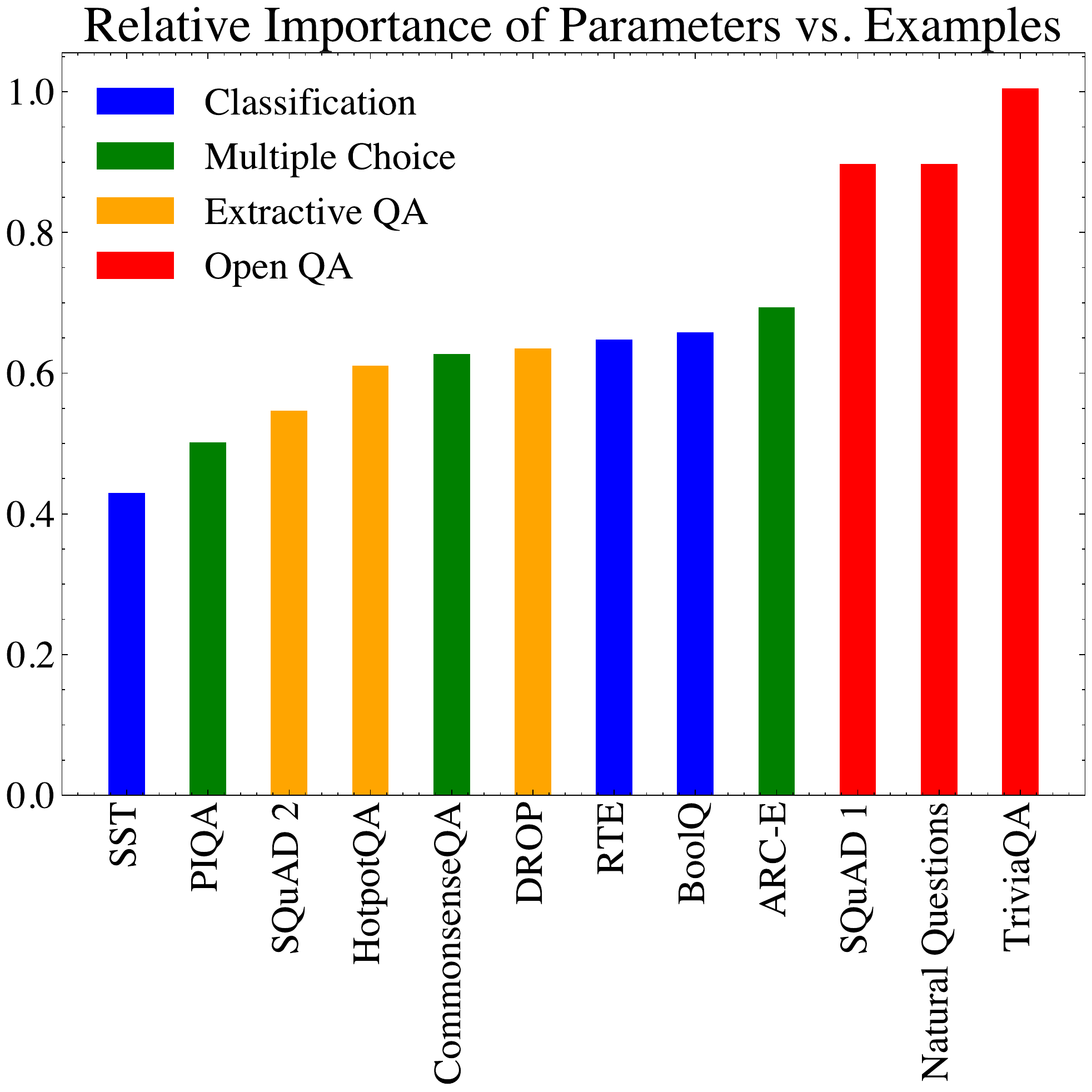}
\caption{The relative importance of parameters versus examples ($I_m$), as computed via regression over each tasks's heatmap. Higher values indicate more dependence on parameters and less on labeled data.}
\label{fig:params_vs_examples}
\end{figure}

\Cref{fig:params_vs_examples} shows that most tasks lie between $0.4 < I_m < 0.7$, with model parameters responsible for most performance improvements, but with significant improvements arising from labeled data as well.
However, all open QA tasks deviate from this interval, and exhibit $I_m$ values of $0.9$ and above, indicating that increased model parameters is almost exclusively responsible for better performance.

\subsection{Comparison with Massive Models}
\label{subsec:comparison}

While models can benefit from both parameters and labeled data in many tasks, scaling up language models to hundreds of billions of parameters may restrict the ability to fine-tune, as GPT3-scale models are typically available only as a service to most practitioners and researchers~\cite{brown2020language}.
Given this data-parameter trade-off, how many labeled examples are 175B parameters worth?

We compare our results of T5-L (800M parameters) fine-tuned on various dataset sizes to those of GPT3 (175B, over 200 times larger than T5-L) using in-context learning, as reported by \citet{brown2020language}.\footnote{Our work has 6 datasets that properly overlap with the original GPT3 paper. ARC-E and PIQA are also used to evaluate GPT3, but in the open QA format.}
Table~\ref{tab:gpt3} shows a wide performance gap between GPT3 and T5-L on open QA datasets, which cannot be bridged by additional labeled examples, as observed in our main experiments.
However, for classification and extractive QA tasks, even a few hundred labeled examples are often enough for T5-L to catch up with GPT3's performance and even exceed it.
In BoolQ, for example, just collecting 96 additional examples is tantamount to adding 200 times more parameters to the model.
This result demonstrates that while performance may  improve along the parameter axis in classification and extractive QA tasks, a small amount of labeled training data can also go a long way.

\begin{table*}[t]
\centering
\small
\begin{tabular}{@{}lrrrrrrr@{}}
\toprule
\multirow{2}{*}{\textbf{Model}} & \multirow{2}{*}{\textbf{\#Examples}} & \multicolumn{2}{c}{\textbf{Classification}} & \multicolumn{2}{c}{\textbf{Extractive QA}} & \multicolumn{2}{c}{\textbf{Open QA}} \\
& & RTE & BoolQ & SQuAD 2 & DROP & NQ & TriviaQA \\
\midrule
\multirow{4}{*}{\textbf{T5-L}}
& 32 & 68.2 & 65.0 & 39.4 & 23.8 & 5.1 & 12.0 \\
& 128 & \textbf{72.9} & 77.0 & 55.3 & 26.5 & 5.8 & 11.5 \\
& 512 & \textbf{79.1} & \textbf{82.5} & \textbf{70.5} & 31.2 & 5.6 & 12.1 \\
& 2048 & \textbf{86.3} & \textbf{85.4} & \textbf{78.9} & \textbf{37.7} & 6.3 & 10.7 \\
\midrule
\textbf{GPT3} 
& $\le$64 & 72.9 & 77.5 & 69.8 & 36.5 & 29.9 & 71.2 \\
\bottomrule
\end{tabular}
\caption{A comparison between GPT3 (with in-context learning, as reported by \citet{brown2020language}) and T5-L. Figures in bold represent T5-L configurations that outperform GPT3. For a fair comparison with \citet{brown2020language}, we report accuracy (exact match) for open QA tasks in this table.}
\label{tab:gpt3}
\end{table*}

\subsection{Discussion}
\label{subsec:discussion}

Why does changing the task's input format have such a dramatic effect on the training dynamics?
We conjecture that the format changes described in our experiments, which effectively remove information from the input, force the models to supplement that information with knowledge stored in its parameters. 
For example, when asking \textit{What is the capital of Micronesia?} in the open QA format, the model is required to know that the answer is \textit{Palikir} by encountering the fact during pretraining or fine-tuning on a paraphrase of the same question.
In contrast, if the same question is asked in the multiple choice format, and the options are \textit{(1) Rome, (2) Tokyo, (3) Yaren, (4) Palikir}, the model can easily eliminate the more frequently-mentioned capitals of Rome and Tokyo, and then guess between the two remaining options, \textit{Yaren} (the capital of neighboring Nauru) and \textit{Palikir} (the correct answer).
A similar example can be constructed for extractive QA, where the vast majority of passage spans can be pruned a priori, leaving only a handful of named entities as more likely candidates.
We hypothesize that answering strategies, such as elimination, can indeed be learned from small-medium training sets, while actual new facts cannot, unless there is significant train-test overlap \cite{lewis-etal-2021-question}.

A practical corollary of this hypothesis is that if one can modify a target task from the open QA format to one with a more limited output space, à la multiple choice or extractive QA, they would unlock the ability to trade data for parameters.
Instead of relying on massive pretrained language models, which can only be used as a service, one could achieve competitive and even superior results with a much smaller model, given a relatively small dataset of several hundred labeled examples.
Retrieve-and-read approaches successfully demonstrate this notion by decomposing open QA into two separate classification and extractive QA subproblems~\cite{chen-etal-2017-reading, lee-etal-2019-latent, karpukhin-etal-2020-dense}, and may possibly be applied to few-shot scenarios in additional tasks via more general retrieve-and-generate models such as RAG \cite{rag}.

% Recently, general-purpose retrieval-augmented models, such as REALM~\cite{realm} and RAG \cite{rag} have been proposed.
% We speculate that such approaches could show promise in few-shot learning contexts, if they can be equipped with sufficiently high-quality and general-purpose retrieval components which can surface inputs for which fast-to-learn solving strategies can be applied.

\section{Related Work}
% \patrick{
% Few-shot learning has been a subject of interest for several decades, often approached via the lens of metric- and meta-learning~\cite{thrun_learning},
% %
% with early examples tending to focus on computer vision tasks~\cite{fink_object,FeiFei2006OneshotLO}. 
% %
% Some metric-learning examples for NLP tasks include \citet{matchingnetworks} and \citet{yu-etal-2018-diverse}, with meta-learning-based approaches often based on MAML~\cite{maml,fewshot_maml}.
% % with pre-trained language models arguably falling into the former category,
% Few-shot learning in NLP has traditionally focused on quickly learning new classes and domains within the context of a single task. The reader is referred to \citet{yin_meta} for further detail. 
% }

Few-shot learning has been a subject of interest for several decades~\cite{thrun_learning,fink_object,FeiFei2006OneshotLO,matchingnetworks,fewshot_maml}. Within NLP, ``few-shot learning" has traditionally focused on quickly learning new classes and domains within the context of a single task (see \citet{yin_meta} for a recent survey). 
Recently, there has been a surge of interest in few-shot learning, following the release of GPT3~\cite{brown2020language}. 
Here, the few-shot learning paradigm has shifted subtly, and refers to building models capable of tackling a range of standard NLP tasks, albeit using very restricted training sets, usually sub-sampled from the full training set.
A great deal of work has recently been produced in this area, and we provide a necessarily incomplete summary below.

\paragraph{In-Context Learning} In-context learning~\cite{brown2020language} generally refers to adapting to a task by providing training examples as additional textual input, without performing gradient-based updates. This technique 
% was used by \citet{brown2020language}, and can be more efficient than fine-tuning, 
 imposes a limit on size of the training dataset due to context length limits. Recent work from \citet{Liu2021WhatMG} and \citet{fantastic_prompts} demonstrate that the choice of in-context training examples, and the order in which they are presented have large effects on performance.

\paragraph{Prompt-Based Learning} Prompting refers to providing additional input to a model designed to help it to produce correct outputs. Typically, these take the form of textual templates used to form cloze questions, and have been used in a variety of settings, such as probing~\cite{petroni-etal-2019-language} and zero-shot learning~\cite{radford2019language}. Prompts can be used in conjunction with fine-tuning, which has been shown to improve results  in a number of works~\cite{Schick2020FewShotTG,Schick2021ItsNJ,schick-schutze-2021-exploiting,gao-etal-2021-making,le-scao-rush-2021-many,Tam2021ImprovingAS}. We adopt this technique in our experiments and adapt the models using prompt-based fine-tuning. 

\paragraph{Prompt Engineering} Models may be sensitive to the choice of prompt (especially without fine-tuning), and a number of works attempt to optimize the prompt for the task at hand~\cite{jiang-etal-2020-know,shin-etal-2020-autoprompt}. Recently, a number of works have also proposed generalizing prompts to include task-specific parameters and embeddings, typically learnt via gradient descent while keeping parts or all of the model's parameters frozen~\cite{adaptors,ptuning,optiprompt,qin_learning_how_to_ask,li-liang-2021-prefix,prompt_tuning, RobertLLogan2021CuttingDO}. While these techniques can improve results for frozen models, they generally do not outperform fine-tuning the whole model~\cite{prompt_tuning}, hence we choose to focus on full-model finetuning with standard prompts in our experiments.

\paragraph{Few-Shot Learning Analysis} Closest to our contribution are works placing an  emphasis on the analysis of few-shot model behaviour, rather than focusing on schemes to improve performance. \citet{le-scao-rush-2021-many} quantify the benefit of prompting in few-shot learning, and \citet{perez_true} critically discuss the difficulty of model selection and very low dataset sizes in few-shot learning. Our work is complementary, exploring the relationship between scale, dataset size, and task open-endedness.

\paragraph{Task Formats}
Another important aspect of our work is the investigation of learning as a function of task format. Related work in this area includes research investigating reformulating a task into a different format, such as reducing tasks to NLI~\cite{white-etal-2017-inference, wang-etal-2018-glue} or reading comprehension~\cite{levy-etal-2017-zero,wu-etal-2020-corefqa}, or even reducing all tasks to a single format~\cite{ama,McCann2018TheNL}. A related line of work seeks to understand tasks and datasets by changing or removing parts of the input, and, in-so-doing, changing the task format. Examples include hypothesis-only NLI baselines~\cite{gururangan-etal-2018-annotation, poliak-etal-2018-hypothesis},
and document-only baselines in Reading Comprehension~\cite{kaushik-lipton-2018-much,Sugawara_Stenetorp_Inui_Aizawa_2020}. We also change the available input to a model for a given task, effectively changing the task format, while keeping the targets unchanged. We do this to measure the effect of the open-endedness of a task on sample complexity for differently sized models.

\section{Conclusions}

In this work, we present an empirical investigation on the relationships between (1) a task's format, (2) the number of labeled examples available for said task, and (3) the number of parameters the model tackling the task has.
Through our extensive experiments, we determine that task format greatly affects the relative performance improvement that can be expected from increased training set size and parameter count.
For tasks that do not require the recollection of specific external information -- i.e. classification, multiple choice, and extractive QA -- we find that more labeled data and larger models both reliably improve performance.
In fact, for some of these tasks, adding a few hundred labeled examples is \emph{more} beneficial than scaling up the model size by billions of parameters.
It seems then, from a practitioner's perspective, that for many tasks where data is very sparse, the tried-and-true strategy of simply collecting more training data will often be a more effective strategy than attempting to scale to larger, more computationally-demanding models.
However, the picture is very different for open QA tasks; for such tasks, we find that increasing the size of the training data barely improves performance, leaving parameter inflation as the only reliable approach to improve accuracy.
Finally, we provide a hypothesis to explain these results and conclude with a practical corollary -- when possible, changing the format from open QA into a more ``self-contained'' one will allow labeled data to bridge performance gaps between moderately-sized models and much larger ones.

\section*{Acknowledgments}
We thank Avia Efrat and Ori Ram for valuable feedback and discussions.

\bibliographystyle{acl_natbib}
\bibliography{anthology,acl2021}

\appendix
\clearpage
\section{Tabular Results}
\label{app:full_results}

\Cref{tab:full_results_app} provides the results from our main experiment (\Cref{sec:results}, \Cref{fig:main_heatmaps}) in tabular form.

\begin{table*}[b]
\centering
\scriptsize
\begin{tabular}{@{}lrrrrrrrrrrrrr@{}}
\toprule
\multirow{2}{*}{\textbf{Model}} & \multirow{2}{*}{\textbf{\#Examples}} & \multicolumn{3}{c}{\textbf{Classification}} & \multicolumn{3}{c}{\textbf{Extractive QA}}  & \multicolumn{3}{c}{\textbf{Multiple Choice}} & \multicolumn{3}{c}{\textbf{Open QA}} \\
& & RTE & SST & BoolQ & SQuAD2 & HPQA & DROP & ARC-E & PIQA & CSQA & NQs & TQA & SQuAD1 \\
\midrule
\multirow{4}{*}{\textbf{T5-S}}
& 32 & 53.8 & 65.4 & 57.3 & 37.2 & 28.1 & 6.0 & 29.1 & 55.2 & 21.4 & 1.9 & 4.8 & 3.3 \\ & 128 & 50.9 & 81.5 & 58.0 & 40.5 & 36.5 & 9.1 & 31.5 & 52.9 & 22.4 & 4.5 & 5.8 & 3.9 \\ & 512 & 52.0 & 84.1 & 58.3 & 42.5 & 52.1 & 15.1 & 32.1 & 55.1 & 28.5 & 3.9 & 5.8 & 3.5 \\ & 2048 & 59.6 & 88.0 & 64.9 & 51.1 & 61.8 & 18.9 & 40.7 & 57.3 & 36.0 & 4.4 & 5.7 & 5.0 \\ \midrule
\multirow{4}{*}{\textbf{T5-B}}
& 32 & 53.1 & 73.0 & 58.8 & 34.7 & 48.3 & 10.2 & 31.4 & 52.1 & 26.0 & 5.3 & 10.3 & 7.9 \\ & 128 & 55.1 & 91.9 & 58.5 & 51.1 & 58.4 & 11.8 & 33.0 & 54.9 & 21.8 & 7.8 & 9.7 & 6.7 \\ & 512 & 63.5 & 90.5 & 71.2 & 49.6 & 71.1 & 23.2 & 34.8 & 55.0 & 42.0 & 7.7 & 11.0 & 8.5 \\ & 2048 & 67.5 & 92.9 & 70.4 & 59.8 & 72.7 & 24.5 & 47.7 & 59.6 & 54.5 & 7.5 & 9.6 & 9.0 \\ \midrule
\multirow{4}{*}{\textbf{T5-L}}
& 32 & 68.2 & 84.5 & 65.0 & 39.4 & 71.1 & 23.8 & 42.9 & 55.6 & 41.7 & 10.6 & 17.7 & 11.1 \\ & 128 & 72.9 & 92.4 & 77.0 & 55.3 & 74.9 & 26.5 & 55.2 & 56.6 & 57.8 & 11.8 & 17.4 & 12.0 \\ & 512 & 79.1 & 94.2 & 82.5 & 70.5 & 78.2 & 31.2 & 65.1 & 68.6 & 64.3 & 11.6 & 18.7 & 12.6 \\ & 2048 & 86.3 & 95.1 & 85.4 & 78.9 & 79.7 & 37.7 & 72.6 & 72.6 & 72.7 & 12.0 & 17.1 & 13.0 \\ \midrule
\multirow{4}{*}{\textbf{T5-XL}}
& 32 & 63.7 & 82.3 & 76.5 & 56.4 & 67.4 & 30.1 & 61.4 & 56.2 & 52.4 & 15.4 & 26.2 & 15.4 \\ & 128 & 83.4 & 91.5 & 83.0 & 73.2 & 76.9 & 32.5 & 72.2 & 59.2 & 64.8 & 16.4 & 27.3 & 15.0 \\ & 512 & 84.1 & 95.2 & 85.9 & 77.7 & 80.2 & 36.8 & 76.9 & 75.5 & 74.0 & 16.1 & 27.5 & 15.8 \\ & 2048 & 88.4 & 96.0 & 88.0 & 84.8 & 81.9 & 43.6 & 80.9 & 81.1 & 78.5 & 16.0 & 25.2 & 15.7 \\ 
\bottomrule
\end{tabular}
\caption{The performance (F1/accuracy) of different T5 models fine-tuned on different training set sizes, across 12 different datasets.}
\label{tab:full_results_app}
\end{table*}

\end{document}